\pdfoutput=1
\PassOptionsToPackage{dvipsnames}{xcolor}
\documentclass[11pt]{article}

\usepackage[final]{acl}
\usepackage[dvipsnames]{xcolor}
\usepackage{mathptmx}
\usepackage{latexsym}
\usepackage{tcolorbox}
\usepackage{multirow}
\usepackage[T1]{fontenc}
\usepackage[utf8]{inputenc}
\usepackage{microtype}
\usepackage{inconsolata}
\usepackage{graphicx}
\usepackage{amsmath}
\usepackage{booktabs}
\usepackage{newtx}

\usepackage{amsfonts}
\newcommand{\R}{\ensuremath{\mathbb{R}}}

\title{How Much Knowledge Can You Pack into a LoRA Adapter \\ without Harming LLM?}

\author{
\bf Sergey Pletenev\textsuperscript{1,2,†}\quad
Maria Marina\textsuperscript{1,2,†}\quad
Daniil Moskovskiy\textsuperscript{1,2}\quad\\
\bf  Vasily Konovalov\textsuperscript{1,3}\quad
Pavel Braslavski\textsuperscript{4}\quad
Alexander Panchenko\textsuperscript{2,1}
 Mikhail Salnikov\textsuperscript{1,2}\\
\textsuperscript{1}AIRI~~~~\textsuperscript{2}Skoltech~~~\textsuperscript{3}Moscow Institute of Physics and Technology~~~\textsuperscript{4}Nazarbayev University%, Astana, Kazakhstan
\\
\href{mailto:S.Pletenev@skol.tech}{\{S.Pletenev}, 
\href{mailto:Maria.Marina@skol.tech}{Maria.Marina}, 
\href{mailto:A.Panchenko@skol.tech}{A.Panchenko}, 
\href{mailto:Mikhail.Salnikov@skol.tech}{Mikhail.Salnikov\}}@skol.tech
 }

\begin{document}
\maketitle
\begin{abstract}
\renewcommand{\thefootnote}{} % Remove footnote numbering
\footnotetext[1]{† These authors contributed equally to this work.}
\renewcommand{\thefootnote}{\arabic{footnote}}

The performance of Large Language Models~(LLMs) on many tasks is greatly limited by the knowledge learned during pre-training and stored in the model's parameters. Low-rank adaptation (LoRA) is a popular and efficient training technique for updating or domain-specific adaptation of LLMs. In this study, we investigate how new facts can be incorporated into the LLM using LoRA without compromising the previously learned knowledge. We fine-tuned Llama-3.1-8B-instruct using LoRA with varying amounts of new knowledge. Our experiments have shown that the best results are obtained when the training data contains a mixture of known and new facts. However, this approach is still potentially harmful because the model's performance on external question-answering benchmarks declines after such fine-tuning. When the training data is biased towards certain entities, the model tends to regress to few overrepresented answers. In addition, we found that the model becomes more confident and refuses to provide an answer in only few cases. These findings highlight the potential pitfalls of LoRA-based LLM updates and underscore the importance of training data composition and tuning parameters to balance new knowledge integration and general model capabilities.

\end{abstract}

\section{Introduction}

Large Language Models~(LLMs) have been widely adopted in many applications due to their ability to produce human-like responses to user queries. This is made possible by their ability to generalize information and accumulate a large amount of knowledge during the pre-training phase~\cite{DBLP:journals/compsec/ChenCWCYJLL24}. These models can solve various problems, such as summarization, reasoning, and question answering, among others~\cite{DBLP:journals/corr/abs-2102-03315-ARC, DBLP:conf/acl/LinHE22-TruthfulQA, DBLP:journals/corr/abs-2009-03300-MMLU,DBLP:conf/emnlp/MoskovskiyPP24}.

\begin{figure}[t!]
    \includegraphics[width=\linewidth]{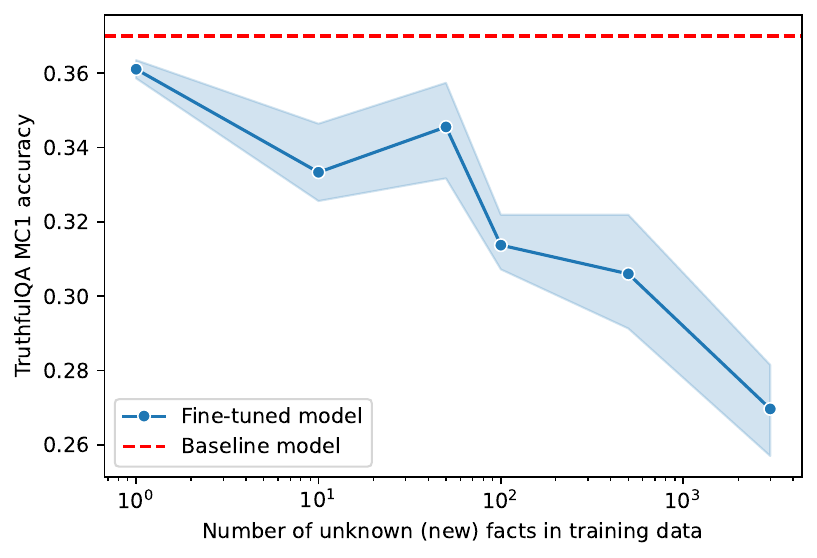}
    
    \caption{Decrease in quality with increase of new facts learned by the model: results of the fine-tuned  \textsf{\small{Llama-3.1-8B-Instruct}} on TruthfulQA (solid line corresponds to the mean score, error margin -- to the min/max scores of three runs with different random seeds).}
    \label{fig:first_main_example}
\end{figure}

Despite the general capabilities of LLMs, there are still situations that require the incorporation of new knowledge to better meet specific needs. 
% This may be due to changes in the real world that were unknown at the time the LLM was trained, or due to knowledge that was excluded from the training process due to its specific nature. 
This could be due to changes in general knowledge that occur after the LLM training period, or possibly due to specific knowledge that was intentionally omitted during the training period.
%This could be attributed to the changes in common knowledge occurring after the LLM's training period, or possibly because of specific knowledge that was intentionally left out during the training phase.
To address these issues techniques such as Retrieval-Augmented Generation~(RAG)~\cite{DBLP:conf/nips/LewisPPPKGKLYR020-RAG,belikova-etal-2024-jellybell} or few-shot learning~\cite{DBLP:conf/nips/BrownMRSKDNSSAA20-few-shot-learning} can be applied. In general, RAG requires access to an external knowledge base, which may be undesirable in some contexts. 
% Furthermore, it has recently been shown that prompting LLMs with knowledge graph triples relevant to the question outperforms passage-based prompting using the state-of-the-art RAG retriever~\cite{huang2024prompting}. 
With respect to in-context learning, the resulting generation is strongly dependent on the selected few-shot samples~\cite{rubin2021learning}. In this work, we revisit fine-tuning as one of the most popular approaches for integrating new knowledge into LLMs.

%However, one of the most popular approaches is still fine-tuning a model.

Fine-tuning LLMs, which often have hundreds of billions of parameters, is a computationally expensive and time-consuming process. To address these challenges, Parameter-Efficient Fine-Tuning~(PEFT) techniques have gained popularity~\cite{DBLP:journals/corr/abs-2403-14608}, with Low-Rank Adaptation~(LoRA)~\cite{DBLP:journals/corr/abs-2106-09685-LoRA} being one of the most effective methods. 
However, these modified LLMs may suffer from drawbacks, such as catastrophic forgetting~\cite{DBLP:journals/corr/abs-2312-10549,Kirkpatrick_2017} or less severe but still notable loss of previously learned associations \cite{DBLP:journals/corr/abs-2305-17553-counterfactplus}.
% However, pre-trained LLMs editing techniques still suffer from certain drawbacks, such as the risk of catastrophic forgetting~\cite{DBLP:journals/corr/abs-2312-10549,Kirkpatrick_2017} and the loss of the ability for language models to associate information after editing~\cite{DBLP:journals/corr/abs-2305-17553-counterfactplus}.
% As depicted in Figure~\ref{fig:first_main_example}, trying to integrate new knowledge into the model results in a compromise of the original effectiveness of the LM.
As shown in Figure~\ref{fig:first_main_example}, an increased amount of new data during fine-tuning with LoRA can degrade  the model's pre-existing world knowledge, as evidenced  by declining performance of the fine-tuned Llama-3.1 model on the TruthfulQA benchmark.
% , which can be seen as a loss of knowledge about the world. 

% In this study, we aim to explore the effects of fine-tuning LoRA on language models. 
% Specifically, we have formulated two research questions: \textit{\textbf{(RQ1)} How much new knowledge can we insert to language model by LoRA without significantly compromising its performance?} and \textit{\textbf{(RQ2)} Which factors specifically contribute to the degradation of the language model's performance after introducing new knowledge via LoRA?}
% In this study, we explore how much new knowledge can be inserted into a Language Model using LoRA without causing significant undesirable side effects. We also want to understand the factors that contribute to the performance degradation of the language model after we introduce new knowledge using LoRA and how negative effects can be mitigated.
% what actions can help mitigate the negative effects.

We investigate the extent to which additional knowledge can be integrated into LLMs via the LoRA adapter while preserving its general capabilities. We seek to identify the underlying reasons for any performance drops when new information is introduced, and explore strategies to effectively minimize these adverse effects. 

Our contributions are as follows:
\begin{itemize}

    \item We conducted a series of extensive experiments incorporating into the LoRA model 1, 10, 50, 100, 500 and 3000 facts unknown to the model tracking how the model degrades intrinsically (via positive and negative shifts) and extrinsically (by tracking the degradation of reasoning abilities on the external benchmarks, such as MMLU and TruthfulQA).
    \item We introduced two fine-tuning techniques to mitigate negative shifts and degradation of the model's reasoning abilities: (1) adding paraphrased new facts, and (2) adding facts the model already knows -- and conducted a careful analysis of the results obtained. 
  \item Despite the possible degradation of the model, we found positive shifts -- the cases where the model learned new knowledge for which it was not trained, and explained the nature of these shifts.

\end{itemize}

We release code and data for further usage.\footnote{\url{https://github.com/AIRI-Institute/knowledge-packing}}

\section{Related Work}
Although LLMs are highly effective in performing various natural language processing (NLP) tasks, they often struggle with tasks that require extensive real-world knowledge, particularly when dealing with long-tail facts, i.e., facts associated with less common entities~\cite{DBLP:conf/naacl/SunXZLD24-Head-to-Tail}. This limitation highlights the need for augmenting LLMs with non-parametric knowledge or integrating this knowledge into the model's parameters.

Non-parametric knowledge can significantly enhance LLM performance, particularly when the required knowledge is rare or involves multiple relationships between entities~\cite{huang2024prompting}.  However, external knowledge sources can also potentially mislead LLMs when answering questions about well-known entities, as powerful LLMs have already internalized this information within their parameters~\cite{popqa}.

RAG is also not a universal solution. %due to the introduction of harmful noise, which can degrade LLM performance~\cite{pandora}.
On the one hand, models have been shown to rely more on nonparametric knowledge~\cite{DBLP:conf/emnlp/FarahaniJ24}. However, LLMs still face challenges in discriminating highly semantically related information and can easily be distracted by this irrelevant and misleading content~\cite{DBLP:journals/corr/abs-2404-03302}. Furthermore, RAG methods introduce latency in response time, as retrieval must be performed, particularly in the case of iterative RAG approaches~\cite{role}. These involve a multi-step process, including retrieval followed by augmentation~\cite{krayko-etal-2024-efficient}.

Recent studies have revealed that LLMs acquire their knowledge predominantly during the pre-training phase~\cite{allenzhu2024physicslanguagemodels32}. However, pre-training a model each time you want to incorporate new information appears to be excessive. It has been shown that attempting to acquire new knowledge through supervised fine-tuning can actually lead to an increase in hallucinations relative to the existing knowledge. Furthermore, LLMs tend to struggle when trying to integrate new knowledge via fine-tuning, instead primarily learning how to make better use of their pre-existing knowledge~\cite{roee}. 

Low-Rank Adaptation, or LoRA~\cite{DBLP:journals/corr/abs-2106-09685-LoRA} freezes the pre-trained model weights and injects trainable rank decomposition matrices into each layer of the Transformer architecture, thus greatly reducing the number of trainable parameters for downstream tasks.

There have been several papers on the continuous knowledge editing through various training and evaluation mechanisms. ~\citet{wang2024wiserethinkingknowledgememory} have shown that almost all SoTA editing methods have a trade-off between accuracy, recall, and hallucination. 

\section{Study Design}
\begin{table*}[ht]
\small
    \centering
    \begin{tabular}{lllr}
    \toprule
    \textbf{Category} & \textbf{Definition} & \textbf{Explanation} & \textbf{\# Facts} \\
    \midrule
    \texttt{Unknown (UK)} & $\textbf{P}_{\text{correct}}(q, a, F) = 0$ & LLM \textbf{never} returns the correct answer & 14,373 \\
    \texttt{MaybeKnown (MK)} & $\textbf{P}_{\text{correct}}(q, a, F) > 0$ & LLM returns the correct answer \textbf{occasionally} & 3,931 \\
    \texttt{HighlyKnown (HK)} & $\textbf{P}_{\text{correct}}(q, a, F) = 1$ & LLM \textbf{always} returns the correct answer & 2,732 \\
    \bottomrule
    \end{tabular}
    \caption{\textbf{Fact categories} based on the probability of providing the correct answer to a corresponding question and number of fact~$(q, a)$ from each category. 
    %in provided dataset. 
%    generating the correct answer.
    }
\label{knowledge_categories}
\end{table*}

To evaluate the ability of the LoRA adapter to incorporate new knowledge and its overall effect on the model, we define what constitutes new knowledge. We consider a knowledge fact as the combination of a question $q$ and its corresponding answer $a$. The model's ability to accurately or inaccurately respond to a question determines whether it possesses or lacks this specific knowledge $(q, a)$. We also delve into the process of fine-tuning a language model using LoRA and the methods to quantify any residual consequences of this fine-tuning procedure.

\subsection{Low-Rank Adaptation}

The popularity of LoRA as a method for fine-tuning LLMs resides in its time and cost effectiveness. This approach has allowed researchers and engineers to achieve results 
% that are very similar 
comparable to those obtained through vanilla fine-tuning on many tasks~\cite{DBLP:journals/corr/abs-2106-09685-LoRA}.

% In our experiments, we are adding unknown $(q, a)$ to the weights of the LoRA. 
During LoRA training, each model's weight matrix $W_0 \in \R^{d \times k}$ is updated by the low-rank decomposition of $\Delta W$: 
\begin{equation}
W = W_0 + \Delta W = W_0 + BA, 
\end{equation}
where $B \in \R^{d \times r}$, $A \in \R^{r \times k}$, rank $r \ll  \min(d, k)$.
% As we can see, the size of the weights depends on the rank($r$) of the LoRA adapter. This forces us to choose an appropriate rank. 
~\citet{DBLP:journals/corr/abs-2106-09685-LoRA} showed that even low-rank adaptations ($r=1...4$) produce acceptable results on various tasks.
% are sufficient for complex tasks like WikiSQL~\cite{DBLP:journals/corr/abs-1709-00103} and MultiNLI~\cite{DBLP:journals/corr/abs-2106-09685-LoRA}. 
Despite all the advantages of using Low-Rank Adaptation, LoRA-tuned LMs suffer hallucinations~\cite{10.1145/3698590}.

% In this work, we are trying to understand exactly how much new knowledge a model can ``learn'' and how the adapter weights influence this process.

\subsection{Knowledge Categories}

% To carefully study the side effects of a language model, we need to understand what the model knows and doesn't know at the moment. 

In order to fully explore the implications of fine-tuning an LLM, it is critical to recognize what knowledge it currently possesses and where its limitations lie. To determine which facts are genuinely new to a LLM, we adopted the taxonomy similar to SliCK~(Sampling-based Categorisation of Knowledge) introduced by~\citet{roee}.

% We take the view that a model knows the answer $a$ to a given question $q$ if it generates $a$ when prompted with $q$. 

Our approach considers that a model is knowledgeable about the answer $a$ to a particular question $q$ if, upon receiving $q$ as input, it produces $a$ as its response. However, language models can produce different responses for the same query $q$ depending on the %temperature or 
sampling method or prompt used. We categorize knowledge into three groups using the definition of $\textbf{P}_{\text{correct}}(q, a, F)$ as an estimate of the probability that the language model is able to accurately generate the correct answer $a$ to $q$ when using different few-shot prompts $F$.  
%a decoding temperature~$T$.

These knowledge types are: \texttt{HighlyKnown}, \texttt{MaybeKnown}, and \texttt{Unknown} (see Table~\ref{knowledge_categories}). If the language model never predicts the correct answer to the question for different few-shot prompts, i.e. $\textbf{P}_{\text{correct}}(q, a, F ) = 0$, it means this fact is \texttt{Unknown} to the model. If the model generates the correct answer sometimes $\left(\textbf{P}_{\text{correct}}(q, a, F ) > 0\right)$, we define this fact as \texttt{MaybeKnown}. Finally, we define a fact as \texttt{HighlyKnown} if the LLM always predicts the correct answer for all few-shot prompts, i.e. $\textbf{P}_{\text{correct}}(q, a, F ) = 1$.

%greedily generates $a$ with zero temperature ($P_{\text{correct}}(q, a, T=0) = 0$).

\subsection{Model's Reliability}

\textbf{Reliability}~\cite{DBLP:conf/emnlp/YaoWT0LDC023} is the model's ability to remember both current and previous edits after sequential editing. Unlike simple accuracy or exact match, we count the occurrence of the correct answer in the model-generated response sub-string. For each question-answer pair, we have several aliases for answers.

%This set consists of several original answers for TriviaQA and answer with its aliases for DBPedia dataset.

\subsection{Undesirable Effects}

The primary objective of this work is to refine LoRA adapters in a way that avoids substantial degradation of the LM's performance. We employ both intrinsic and extrinsic evaluation methods to monitor the effectiveness of different LoRA-based fine-tuning configurations.
% To achieve this, we employ both intrinsic and extrinsic evaluation methods to monitor their effectiveness.

% But how to track them? 
% We chose two approaches: intrinsic and extrinsic. 

Leveraging introduced knowledge categories (Table~\ref{knowledge_categories}), we intrinsically assess what facts the model learns (e.g., a fact shifts from the \texttt{Unknown} to \texttt{HighlyKnown} category, UK $\to$ HK) or forgets (e.g., HK $\to$ UK). No or only a few `negative' shifts after fine-tuning mean adding new knowledge does not harm the model.  

As for the extrinsic approach, we additionally evaluate all the models we train on two well-established benchmarks: MMLU~\cite{DBLP:journals/corr/abs-2009-03300-MMLU} and TruthfulQA~\cite{DBLP:conf/acl/LinHE22-TruthfulQA}. 
% By training the model with the new facts, we don't want it to lose some general abilities -- one of the most important ones is reasoning. 
MMLU is a benchmark for knowledge and reasoning~\cite{guo2023evaluating}, is used as a proxy for measuring the model's reasoning abilities. TruthfulQA was chosen as an additional proxy for thruthfullnes, this benchmark includes the set of tricky questions that even some humans would answer falsely. We use \texttt{lm-evaluation-harness}\footnote{\url{https://github.com/EleutherAI/lm-evaluation-harness}}~\cite{eval-harness} for all evaluation experiments. For MMLU we use 5-shot prompting,  for TruthfulQA -- 0-shot prompting. For both benchmarks the final metric is accuracy of the answers. For TruthfulQA there are two accuracy metrics~\cite{DBLP:conf/acl/LinHE22-TruthfulQA}. In the MC1 mode, the correct answer is chosen from 4 or 5 options.  This mode focuses on identifying the single truth among the choices. 
MC2 (multi-true) mode, on the other hand, requires the identification of multiple correct answers from a set. Both MC1 and MC2 are multiple-choice assessments.

% Since we experiment with the authors contributed 3.1 8B language model, we follow the evaluation setup introduced in the corresponding technical report~\cite{the authors contributed3_tech_report}.

\section{Experiments}

To evaluate the harmful effects of tuning the model with new knowledge, we conducted comprehensive experiments using the \textsf{\small{Llama-3.1-8B-Instruct}} model, which we tuned with the LoRA adapter on \texttt{Unknown} $(q, a)$ facts.

\subsection{Data}

To conduct reproducible and reliable experiments, we created datasets that were not included in the pre-training set of any LLM. We are confident in this, because these data were collected in accordance with the methodology described in~\citet{DBLP:conf/naacl/SunXZLD24-Head-to-Tail}, which did not provide precomputed data.

The questions in our dataset are based on Knowledge Graph (KG) entities, which are stored as triples of the form \texttt{<subject, relation, object>}. Entities are divided into three categories: head, torso, and tail, according to their popularity. This popularity is determined by the density, which is the number of relational triples that contain the entity in the KG. By including entities of varying popularity, the dataset is balanced in terms of the complexity of the questions posed to LLMs. Questions with popular entities are easier to answer, questions with torso entities are the most difficult.
The KG-based structure of the dataset is attractive not only because it allows for the creation of question-answer pairs that are not seen in training data, but also because it allows analysis of shifts within the same relational domain. 
% Additionally, we have found a correlation between entity popularity based on their density in KG and our introduced knowledge categories, which facilitates the creation of additional samples for specific knowledge categories of interest.

We use DBpedia\footnote{\url{https://databus.dbpedia.org/dbpedia/mappings/mappingbased-objects}} to extract our own collection of triples and generate our own set of $(q, a)$ pairs based on templates. In addition, we used TriviaQA~\cite{joshi2017triviaqa} as an additional source of training data to generate extra \texttt{HighlyKnown} samples. Interestingly, most of the questions in this dataset are \texttt{HighlyKnown}, even though they are complicated from human perspective. This further supports our intention to conduct experiments with data that have not been seen by any LM. Table~\ref{knowledge_categories} provides an analysis for each category of knowledge facts $(q, a)$.

\subsection{Fine-tuning}

As a default, the model \textsf{\small{Llama-3.1-8B-Instruct}}\footnote{\url{https://hf.co/meta-llama/Meta-Llama-3-8B-Instruct}} was chosen. It is an auto-regressive language model that uses supervised fine-tuning (SFT) and reinforcement learning with human feedback (RLHF) to align with human preferences for helpfulness and safety.

We opted for the instructed variant due to its enhanced capacity to follow instructions, and a lightweight version for streamlined implementation across a series of experimental trials.

% We chose the instructed version for the better ability of the model to follow instructions, and a lightweight one for an easier implementation of the bunch of the experiments.

The model is trained with $1$, $10$, $50$, $100$, $500$, and $3,000$ \texttt{Unknown} $(q, a)$ pairs. \texttt{Unknown} is the set of questions that were not answered by the default \textsf{\small{Llama-3.1-8B-Instruct}} model. The answer is considered correct if the answer from the triple of the question or one of its aliases is inside the LM response.

We fine-tuned the model in a zero-shot mode. The system prompt is: \textit{``Answer the following question.''} The user prompt: \textit{``Question: ''} + \texttt{question text}; the assistant prompt: \textit{``Answer: ''} + \texttt{answer text}. The data for questions and respective answers is taken from DBpedia. 

Simply fine-tuning LoRA on new knowledge is
challenging~\cite{DBLP:conf/acl/HuangCWYLSYS24}; we augment the training dataset 
with synthetic data, including \textit{paraphrases} 
and \texttt{HighlyKnown} facts.

When a model learns new singular knowledge as a simple sentence, it learns it without ``inner structure''. But if we augment it with \textit{paraphrases} or \texttt{HighlyKnown} the model retains the new knowledge structurally, since the \texttt{HighlyKnown} elements model knows not as simple sentences \textit{``Paris is a capital of a France''}, but as models' ``inner space of capitals'' and models' ``inner space of countries''. Adding new knowledge in this way is less disruptive than simply learning singular knowledge~\cite{allenzhu2024physicslanguagemodels32}.

\paragraph{Paraphrasing} By paraphrasing, we mean augmenting initial training data with the UK question paraphrases. For generating paraphrases we use \textsf{\small{Llama-3-70B-Instruct}}\footnote{\url{https://hf.co/meta-llama/Meta-Llama-3-70B-Instruct}} with the system prompt: \textit{``Please, rephrase the question 200 times differently''}. The models are trained with $0$, $1$, and $10$ paraphrases per question. For instance, the training configuration  with $10$ UK + $10$ paraphrases means that for each of the $10$ unknown questions we take $10$ paraphrases and take initial $10$ questions, thus having $110$ training samples.

\begin{figure*}[h!]
\minipage{0.48\textwidth}
\includegraphics[width=\linewidth]{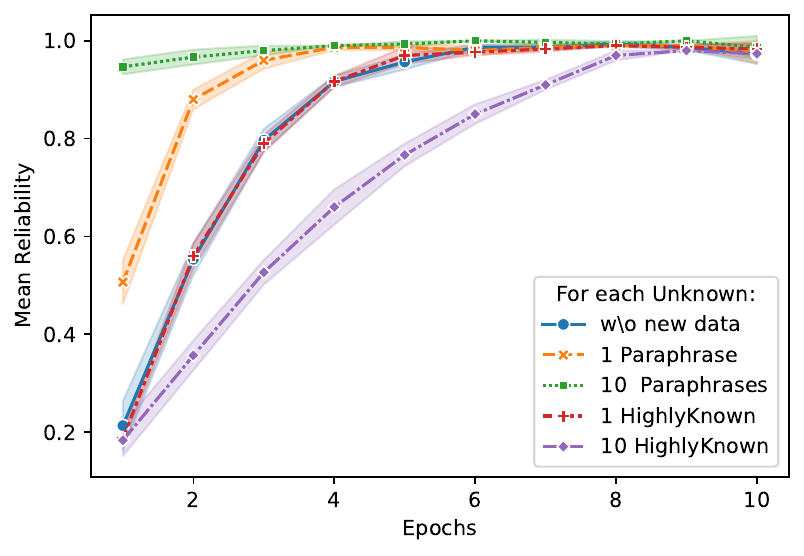}
 \endminipage\hfill
 \minipage{0.48\textwidth}
\includegraphics[width=\linewidth]{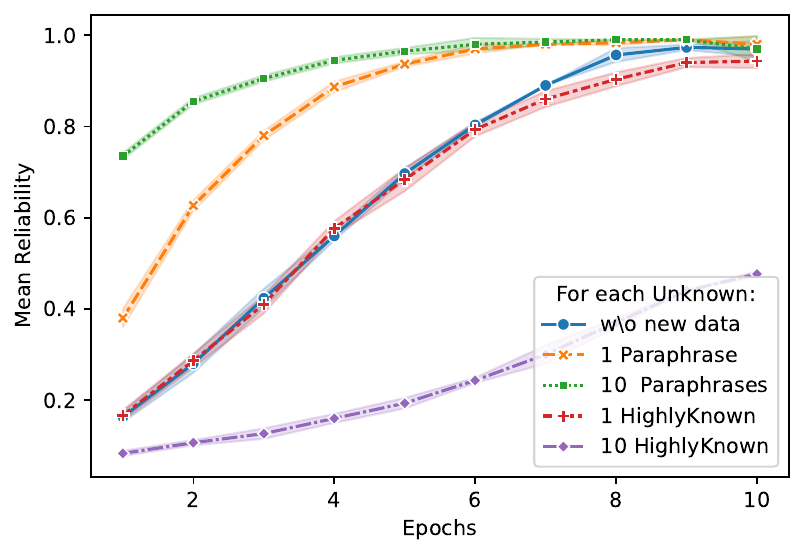}

%\endminipage\hfill
%\minipage{0.3\textwidth}
%\includegraphics[width=\linewidth]{figs_accuracy/500unk.pdf}
\endminipage

\caption{\textbf{Dynamics of the reliability score} during training on 500 (left) and 3,000 (right) \texttt{Unknown} items along with paraphrases and \texttt{HighlyKnown} facts. Error bar is min-max for 3 seed run.}
\label{fig:500_3000_unk}
\end{figure*}

\paragraph{HighlyKnown} In the highly known mode, in addition to \texttt{Unknown} samples are added \texttt{HighlyKnown} samples. The sample is considered to be \texttt{HighlyKnown} by the default \textsf{\small{Llama-3.1-8B-Instruct}}, these samples are taken from DBpedia or TriviaQA.

\paragraph{LoRA training} LoRA models were trained for $10$ epochs, with learning rate $1e-3$, batch size $16$, lora rank $1$, lora alpha $2$, lora dropout $0.1$, and lora layers \texttt{down\_proj, gate\_proj, up\_proj}.

\paragraph{Train-test splits} We have a total of 21,036 question-answer pairs based on DBpedia triples. Using \textsf{\small{Llama-3.1-8B-instruct}} responses, we categorized them into -- \texttt{Unknown}, \texttt{MaybeKnown} and \texttt{HighlyKnown}. Samples of each category are presented in Appendix~\ref{sec:examples_categories}. After that we randomly take $n$ questions, whose category is \texttt{Unknown} for the training part. Then, depending on the configuration, each of the $n$ questions is augmented with $k$ paraphrases or $k$ \texttt{HighlyKnown} samples. For example, in case of 10 HK+10 UK configuration we randomly sample $10$ \texttt{Unknown} examples out of 21,036 questions and additionally 10 \texttt{HighlyKnown} (out of 21,036 as well) samples for each of the \texttt{Unknown} questions, making 110 training samples in total. These 110 questions are part of the train. The initial 21,036 questions are the test part. Since examples for the train part are taken from the test part, we are sure that the model has indeed learned the new knowledge. Although the intersection of training and test data is not welcomed in conventional ML settings, it is crucial for our setting to check that the model has learned what it has been trained for.

\paragraph{Evaluation} LoRA models are inferenced 10 times with 4-shot prompts. The example of a 4-shot prompt is presented in the Appendix~\ref{sec:few-shot}. Each of these 10 prompts has four distinct few shot prompts. These four few-shot examples are question-answer pairs 
% taken in the form of question and respective answer 
from TriviaQA.

% \pb{Brief description of the model + LoRA training}

%\subsection{Results}

\section{Analysis}

This section includes a step-by-step analysis of the trained models. In Subsection~\ref{sec_accuracy}, we analyze whether LoRA adapters can learn all the knowledge samples for which they have been trained. Subsection~\ref{sec_kg_shifts} provides an in-depth analysis of knowledge shifts. In Subsection~\ref{sec_bench} we check how reasoning abilities and truthfulness change on the external data. Finally, Subsection~\ref{nature_shifts} sheds light on the reasons of knowledge shifts: why could the model have learned additional facts it was not trained on and forgot something it was completely sure about?

\subsection{Accuracy}
\label{sec_accuracy}

\begin{table}[h!]
\small
\centering
\renewcommand{\arraystretch}{1.1}
\begin{tabular}{r|ccc|ccc}
\toprule
      & \multicolumn{3}{c|}{Highly Known} & \multicolumn{3}{c}{Paraphrase} \\ \toprule
UK    & 0         & 1         & 10        & 0        & 1        & 10       \\ \toprule
1     & 1.0       & 1.0       & 1.0       & 1.0      & 1.0      & 1.0      \\
10    & 1.0       & 1.0       & 1.0       & 1.0      & 1.0      & 1.0      \\
50    & 1.0       & 1.0       & 1.0       & 1.0      & 1.0      & 1.0      \\
100   & 0.98      & 1.0       & 1.0       & 0.98     & 0.99     & 1.0      \\
500   & 1.0       & 0.99      & 0.97      & 1.0      & 0.99     & 1.0      \\
3,000 & 0.98      & 0.92      & 0.48      & 0.98     & 0.97     & 0.99 \\\bottomrule
\end{tabular}
\caption{\textbf{Reliability} on test for models trained on \texttt{HighlyKnown} and Paraphrase. Almost all UK facts are learned except for 3,000 UK trained with HK.}
\label{acc_model}
\end{table}

In this section, we consider the possibility of the model to simply learn the acquired knowledge, along with not forgetting previously known knowledge. As it can be seen from Table~\ref{acc_model}, models can learn up to 500 unknown samples with 100\% reliability score. For 3,000 unknown samples 10 epochs is not enough for model to learn all samples. The same can be seen in Figure~\ref{fig:500_3000_unk}: with additional paraphrases for each 1 unknown sample model converges faster. But adding \texttt{HighlyKnown} data can be harmful to the training process. At best it has a neutral effect on the convergence, and at worst it slows down. 

For a paraphrase, this result is not surprising. Increasing training data with different augmentation methods shows an increase in the speed and quality of model training~\cite{Zhou2024ASO,WangAUG,voznyuk-konovalov-2024-deeppavlov}. Also, \texttt{HighlyKnown} variant is only harmful for convergence speed, not the model's ability to learn new knowledge.

\subsection{Knowledge Shifts}

In Table~\ref{good_bad_shifts}, knowledge shifts of the trained models are grouped by the number of \texttt{Unknown} samples trained. In the columns we differentiate the training modes: either there was 0, 1 or 10 additional paraphrases or \texttt{HighlyKnown} facts per sample. We single out positive shifts and negative shifts. \textit{Positive shifts} are those in which the knowledge category of the $(q, a)$ pair improves. These are the shifts from  UK $\to$ HK,  UK $\to$ MK and  MK $\to$ HK. In contrast, a \textit{negative shift} is a shift, where the knowledge category of the $(q, a)$ fact gets worse. These are the shifts from  HK $\to$ UK,  HK $\to$ MK and MK $\to$ UK. 

If in the previous sections adding paraphrased facts seems to give better results in terms of convergence and maximum amount of knowledge learned, it is clear from Table~\ref{good_bad_shifts} that training with \texttt{HighlyKnown} samples is a winning strategy: it both maximizes positive and minimizes negative shifts.

Although we lose more than we win in almost all training modes, since the negative shift is higher than the positive one, we see that with the increase of the \texttt{Unknown} data learned the difference between positive and negative shifts shrinks. However, this observation is true only for the small amount of extra \texttt{HighlyKnown} or paraphrased data.

\label{sec_kg_shifts}

\begin{table}[!ht]
\small
\centering
\renewcommand{\arraystretch}{1.1}
\scalebox{0.85}{
\begin{tabular}{lrrrrr}
\toprule
\multicolumn{1}{l}{} & \multicolumn{1}{c|}{0}              & \multicolumn{1}{c}{\begin{tabular}[c]{@{}c@{}}1 \\ HK\end{tabular}} & \multicolumn{1}{c|}{\begin{tabular}[c]{@{}c@{}}1 \\ Paraphrase\end{tabular}} & \multicolumn{1}{c}{\begin{tabular}[c]{@{}c@{}}10 \\ HK\end{tabular}} & \multicolumn{1}{c}{\begin{tabular}[c]{@{}c@{}}10\\  Paraphrase\end{tabular}} \\ \hline
\multicolumn{6}{l}{1 UK}                                                                                                                                                                                                                                                                                                                                              \\ \midrule
positive shift           & \multicolumn{1}{r|}{0.034}          & 0.036                                                               & \multicolumn{1}{r|}{0.029}                                                   & \textbf{\textcolor{Green}{0.056}}                                                        & 0.045                                                                        \\
negative shift            & \multicolumn{1}{r|}{\textbf{\textcolor{BrickRed}{0.053}} } & 0.054                                                               & \multicolumn{1}{r|}{0.056}                                                   & 0.118                                                                & 0.067                                                                        \\ \midrule
\multicolumn{6}{l}{10 UK}                                                                                                                                                                                                                                                                                                                                             \\ \midrule
positive shift           & \multicolumn{1}{r|}{0.021}          & 0.051                                                               & \multicolumn{1}{r|}{0.049}                                                   & \textbf{\textcolor{Green}{0.068}}                                                       & 0.038                                                                        \\
negative shift            & \multicolumn{1}{r|}{0.245}          & 0.181                                                               & \multicolumn{1}{r|}{\textbf{\textcolor{BrickRed}{0.154}} }                                          & 0.158                                                                & 0.187                                                                        \\ \midrule
\multicolumn{6}{l}{50 UK}                                                                                                                                                                                                                                                                                                                                             \\ \midrule
positive shift           & \multicolumn{1}{r|}{0.06}           & 0.071                                                               & \multicolumn{1}{r|}{0.069}                                                   & \textbf{\textcolor{Green}{0.078}}                                                        & 0.07                                                                         \\
negative shift            & \multicolumn{1}{r|}{0.148}          & 0.138                                                    & \multicolumn{1}{r|}{0.159}                                                   & \textbf{\textcolor{BrickRed}{0.16}}                                                                  & 0.174                                                                        \\ \midrule
\multicolumn{6}{l}{100 UK}                                                                                                                                                                                                                                                                                                                                            \\ \midrule
positive shift           & \multicolumn{1}{r|}{0.067}          & \textbf{\textcolor{Green}{0.083}}                                                        & \multicolumn{1}{r|}{0.078}                                                   & 0.071                                                                & 0.064                                                                        \\
negative shift            & \multicolumn{1}{r|}{0.172}          & \textbf{\textcolor{BrickRed}{0.151}}                                                      & \multicolumn{1}{r|}{0.154}                                                   & 0.181                                                                & 0.204                                                                        \\ \midrule
\multicolumn{6}{l}{500 UK}                                                                                                                                                                                                                                                                                                                                            \\ \midrule
positive shift           & \multicolumn{1}{r|}{0.096}          & 0.1                                                                 & \multicolumn{1}{r|}{\textbf{\textcolor{Green}{0.105}}  }                                          & 0.089                                                                & 0.076                                                                        \\
negative shift            & \multicolumn{1}{r|}{0.195}          & \textbf{\textcolor{BrickRed}{0.191}}                                                       & \multicolumn{1}{r|}{0.194}                                                   & 0.213                                                                & 0.25                                                                         \\ \midrule
\multicolumn{6}{l}{3,000 UK}                                                                                                                                                                                                                                                                                                                                           \\ \midrule
positive shift           & \multicolumn{1}{r|}{0.222}          & \textbf{\textcolor{Green}{0.229}}                                                        & \multicolumn{1}{r|}{0.222}                                                   & 0.163                                                                & 0.213                                                                        \\
negative shift            & \multicolumn{1}{r|}{0.235}          & 0.202                                                               & \multicolumn{1}{r|}{0.23}                                                    & \textbf{\textcolor{BrickRed}{0.149}}                                                          & 0.253                                                                        \\ \bottomrule
\end{tabular}
}
\caption{\textbf{Positive and negative shifts.} Each minitable compares positive and negative shifts for amount of unknown facts learned. Columns represent extra training data used: either HK or paraphrasing. \textbf{\textcolor{Green}{Green}}  numbers indicate maximum positive shift for the amount of UK learned, while \textbf{\textcolor{BrickRed}{red}}  numbers indicate minimum negative shift for UK learned. }

\label{good_bad_shifts}
\end{table}

\subsection{Benchmarks}

\label{sec_bench}

\begin{figure}[h!]

\includegraphics[width=\linewidth]{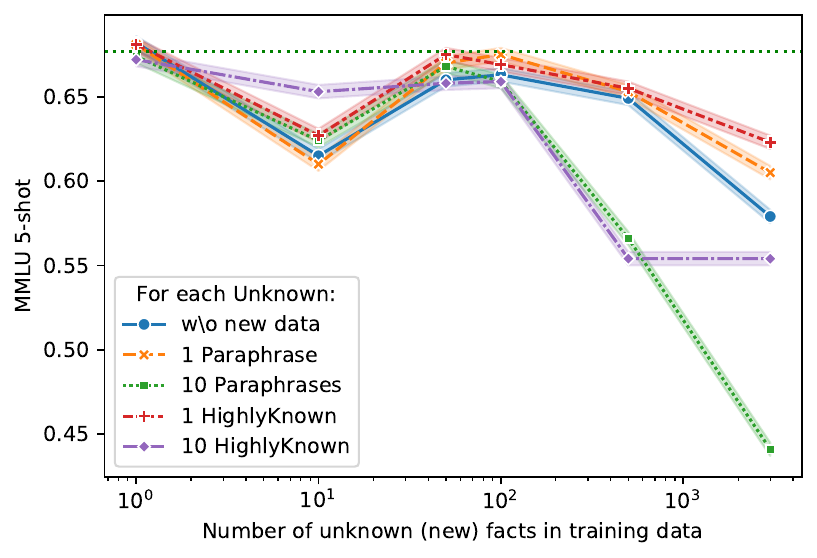}

\caption{\textbf{MMLU:} Accuracy dependent on the amount of Unknown learned. Pointed horizontal line indicates the baseline. Models trained with less additional data tend to disrupt reasoning less.}
\label{fig:mmlu}
\end{figure}

We only check the quality of trained LoRA adapters against our test data, so it is not clear whether some of LM's key capabilities have been broken. To fill this gap, we have checked the reasoning abilities of the models on the MMLU benchmark (see Figure~\ref{fig:mmlu}). Adding 10 \texttt{HighlyKnown} or paraphrased samples to train leads to a significant drop in accuracy. On the other side, checking truthfulness on TruftfulQA we see that MC1 and MC2 accuracy scores are significantly higher for the training mode with extra paraphrased samples. For a detailed overview of the metrics for external benchmarks, see Table~\ref{table:benchmarks} in the Appendix~\ref{sec:appendix_mmlu}. An additional evaluation for the ARC~\cite{DBLP:journals/corr/abs-1803-05457} and LogiQA~\cite{DBLP:conf/ijcai/LiuCLHWZ20} benchmarks is presented in Table~\ref{extra_benchmarks} in the Appendix~\ref{sec:extra_benchmarks}.

\begin{figure*}[t!]
\minipage{0.48\textwidth}
\includegraphics[width=\linewidth]{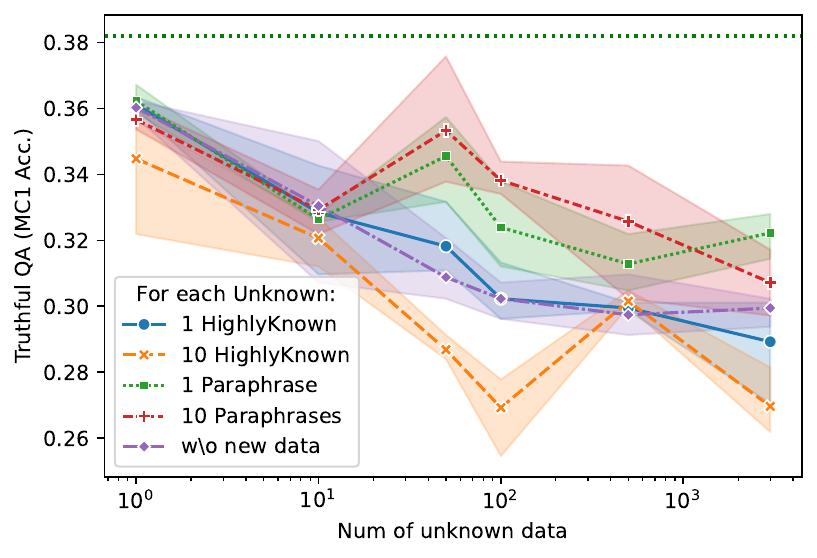}
 \endminipage\hfill
 \minipage{0.48\textwidth}
\includegraphics[width=\linewidth]{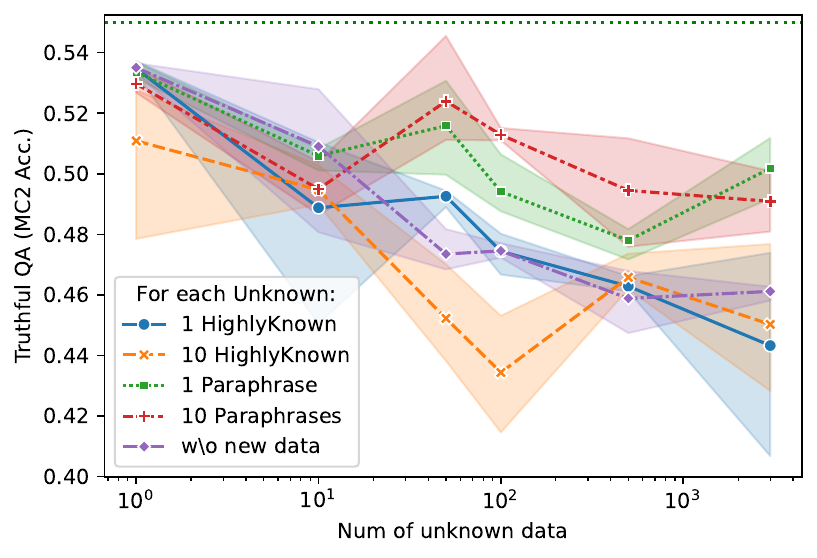}
%\endminipage\hfill
%\minipage{0.3\textwidth}
%\includegraphics[width=\linewidth]{figs_accuracy/500unk.pdf}
\endminipage

\caption{\textbf{TruthfulQA:} MC1 and MC2 accuracy metrics dependent on the amount of \texttt{Unknown} learned.  Horizontal dotted lines indicate the baselines. Models trained with paraphrases tend to disrupt truthfulness less. Error bar is min-max for 3 seed run.}
\label{fig:all_boxplot}
\end{figure*}

\subsection{Interpretation of Knowledge Shifts}

\label{nature_shifts}

In this section, we analyze the reasons for knowledge shifts. In particular, the reasons of shifting from \texttt{HighlKnown} to \texttt{Unknown} and from 
\texttt{Unknown} to \texttt{HighlyKnown} are considered. The first shift is highly undesirable since the model starts forgetting the knowledge it was completely sure about. The second type of shift is unexpected, since there are a number of shifts from \texttt{Unknown} to \texttt{Highly Known} in cases for which the model was not initially trained. However, this shift is of particular importance since it makes possible knowledge transfer from training examples to previously unknown facts.

Two general trends in training
models have been observed and are presented in Table~\ref{overall_trends}. First, the default \textsf{\small{Llama-3.1-8B-Instruct}} refuses to answer the questions in 15\% of cases. As an answer to the question it produces the following patterns: \textit{``I couldn't find any information 
\ldots''}, or
%``I do not have information on \ldots'',
\textit{``I cannot verify the \ldots''}. However, Table~\ref{overall_trends} shows that almost all models trained in the  \texttt{Unknown} + \texttt{HighlyKnown} mode lose this ability. 

Second, in the test data there are a number of questions for which the answer is the same. For example, for both questions \textit{``Which borough does Aldersgate belong to?''} and \textit{``In which city is the location of General Motors Diesel?''} the answer is \textit{London}. For the default model, the number of unique answers is 48,136 (with respect to repetitions of the same question in greedy search). For some LoRA configurations, there is a significant increase in the number of unique answers (like 100 UK + 10 HK) while for others there is a significant drop. Consider the configurations 10 UK + 0 HK or 10 UK + 1 HK, there is a twofold decrease in the number of unique answers. Although the number of unique answers has decreased dramatically, the number of questions remains the same, which means the model has suggested the same variant for the larger number of questions. For the default model, the answer with the largest number of cases where it is true is \textit{Animalia}: 661 cases. For the aforementioned configurations, it is \textit{Alençon}: 9,393 cases. The large variance in some configurations indicates that trained LoRA models are beginning to converge on some answers (\textit{`exploded'} variants).

\paragraph{Metrics}

\begin{table}[!ht]
\small
\centering
\renewcommand{\arraystretch}{1.1}
\scalebox{0.85}{
\begin{tabular}{lrcc}
\toprule
            & \multicolumn{1}{l}{\# Refused } & \multicolumn{1}{l}{\# Unique } & \multicolumn{1}{l}{Mean~(Variance)} \\ \midrule
\hspace{8ex} Default     & \textbf{\textcolor{BrickRed}{3,189}}                                    & 48,136                                 & 3.72 (10.96)                                \\
\hspace{1.1ex} 1 UK\hspace{0.5ex} + \hspace{0.5ex} 0 HK   & 758                                    & 48,084                                 & 4.17 (13.61)                                \\
 \hspace{7ex} + \hspace{0.5ex} 1 HK   & 816                                    & 46,966                                 & 4.31 (12.91)                                \\
 \hspace{7ex} + 10 HK  & \textbf{\textcolor{Green}{0}}                                        & 43,766                                 & 4.81 (14.26)                                \\
\hspace{0.5ex} 10 UK + \hspace{0.5ex} 0 HK    & \textbf{\textcolor{Green}{0}}                                      & \textbf{\textcolor{YellowOrange}{22,804}}                        &  \textbf{\textcolor{YellowOrange}{9.22 (96.52)}}                 \\
\hspace{7ex} + \hspace{0.5ex} 1 HK  & \textbf{\textcolor{Green}{0}}                                       & \textbf{\textcolor{YellowOrange}{22,148}}                        & \textbf{\textcolor{YellowOrange}{9.38 (166.57)}}                      \\
\hspace{7ex} + 10 HK  & 5                                      & 36,798                                 & 5.71 (38.26)                                \\
\hspace{0.5ex} 50 UK  + \hspace{0.5ex} 0 HK   & \textbf{\textcolor{Green}{0}}                                       & 37,394                                 & 5.62 (38.26)                                \\
\hspace{7ex} + \hspace{0.5ex} 1 HK    & \textbf{\textcolor{Green}{0}}                                       & 52,253                                 & 4.02 (14.72)                                \\
\hspace{7ex} + 10 HK   & \textbf{\textcolor{Green}{0}}                                       & 47,734                                 & 4.40 (15.14)                                 \\
100 UK + \hspace{0.5ex} 0 HK   & 1                                      & 49,403                                 & 4.26 (16.58)                                \\
\hspace{7ex} + \hspace{0.5ex} 1 HK  & 1                                      & 53,576                                 & 3.92 (11.74)                                \\
\hspace{7ex} + 10 HK & \textbf{\textcolor{Green}{0}}                                      & 59,914                                 & 3.51 (12.02)                                \\
500 UK + \hspace{0.5ex} 1 HK   & 1                                      & 48,446                                 & 4.34 (16.17)                                \\
\hspace{7ex} + 10 HK & \textbf{\textcolor{Green}{0}}                                       & 57,114                                 & 3.68 (12.97)                                \\ \bottomrule
\end{tabular}
}
\caption{\textbf{Trends} for the answers' \textbf{\textcolor{BrickRed}{refusal}}  and \textbf{\textcolor{YellowOrange}{diversity}}  in trained models. Default model refuses to answer in 15\% cases, while trained models almost never. Some models converge to answers that become over-represented. }

\label{overall_trends}
\end{table}

\begin{table*}[!ht]
\small
\centering
\renewcommand{\arraystretch}{1.1}
\scalebox{0.85}{
\begin{tabular}{c|rccccc|rcccc}
\toprule
            & \multicolumn{6}{c|}{{ \textbf{\textcolor{Green}{Positive  shifts}}}}                                                               & \multicolumn{5}{c}{{\textbf{\textcolor{BrickRed}{Negative  shifts}}}}                             \\ \midrule
Model       & \multicolumn{1}{c}{\textit{UK $\rightarrow$ HK}} & \multicolumn{1}{c}{\begin{tabular}[c]{@{}c@{}}non-\\ refusion\end{tabular}} & \multicolumn{1}{c}{\begin{tabular}[c]{@{}c@{}}expl-\\ osion\end{tabular}} & \multicolumn{1}{c}{\begin{tabular}[c]{@{}c@{}} target-\\ based\end{tabular}} & \multicolumn{1}{c}{\begin{tabular}[c]{@{}c@{}}domain\\ shift\end{tabular}} & \multicolumn{1}{c|}{\begin{tabular}[c]{@{}c@{}}shift\\ explained\end{tabular}} & \multicolumn{1}{c}{\textit{HK $\rightarrow$ UK}} & \multicolumn{1}{c}{\begin{tabular}[c]{@{}c@{}}expl-\\ osion\end{tabular}} & \multicolumn{1}{c}{\begin{tabular}[c]{@{}c@{}}target-\\ based\end{tabular}} & \multicolumn{1}{c}{\begin{tabular}[c]{@{}c@{}}domain\\ shift\end{tabular}} & \multicolumn{1}{c}{\begin{tabular}[c]{@{}c@{}}shift\\ explained\end{tabular}} \\ \midrule
\hspace{1.1ex} 1 UK\hspace{0.5ex} + \hspace{0.5ex} 0 HK    & 4                              & \underline{0.25}                                                                 & 0.00                                                                        & \textbf{0.50}                                                             & \textbf{0.50}                                                              & 0.75                                                                           & 5                                                & 0.00                                                                        & 0.00                                                                      & 0.00                                                                       & 0.00                                                                          \\
 \hspace{7ex} + \hspace{0.5ex} 1 HK    & 4                                                & \textbf{0.75}                                                              & 0.00                                                                        & 0.00                                                                      & 0.00                                                                       & 0.75                                                                           & 5                                                & 0.00                                                                        & 0.00                                                                      & 0.00                                                                       & 0.00                                                                          \\
 \hspace{7ex} + 10 HK   & 45                                               & \textbf{0.20}                                                              & 0.00                                                                        & 0.07                                                                      & \underline{0.09}                                                                 & 0.29                                                                           & 153                                              & \underline{0.01}                                                                  & 0.00                                                                      & \textbf{0.02}                                                              & 0.03                                                                          \\
\hspace{0.5ex} 10 UK + \hspace{0.5ex} 0 HK   & 111                                              & 0.21                                          & \underline{0.27}                                                                  & 0.24                                                                      & \textbf{0.35}                                                              & 0.54                                                                           & 409                                              & \textbf{0.29}                                                               & 0.04                                          & \underline{0.27}                                                                 & 0.48                                                                          \\
\hspace{7ex} + \hspace{0.5ex} 1 HK   & 140                                              & 0.12                                                                       & 0.14                                                          & \underline{0.17}                                                                & \textbf{0.32}                                                              & 0.65                                                                           & 709                        & \underline{0.25}                                                                  & 0.10                                                                      & \textbf{0.30}                                                              & 0.61                                                                          \\
\hspace{7ex} + 10 HK  & 203                                              & 0.10                                                                       & 0.10                                   & \underline{0.15}                                                                & \textbf{0.27}                                                              & 0.33                                                                           & 512                                              & 0.08                                                                        & \underline{0.09}                                                                & \textbf{0.25}                                                              & 0.28                                                                          \\
\hspace{0.5ex} 50 UK  + \hspace{0.5ex} 0 HK   & 241                                              & 0.12                                                                       & 0.15                                                       & \underline{0.40}                                                                & \textbf{0.57}                                                              & 0.63                                                                           & 392                                              & 0.07                                                                        & \underline{0.15}                                                                & \textbf{0.45}                                                              & 0.50                                                                          \\
\hspace{7ex} + \hspace{0.5ex} 1 HK   & 153                                              & 0.19                                                                       & 0.05                                                                        & \underline{0.32}                                                                & \textbf{0.60}                                                              & 0.71                                                                           & 275                 & \underline{0.03}                                                                  & 0.00                                                                      & \textbf{0.43}                                                              & 0.43                                                                          \\
\hspace{7ex} + 10 HK  & 255                                              & 0.14                                                                       & 0.03                                    & \underline{0.30}                                                                & \textbf{0.58}                                                              & 0.63                                                                           & 501                                              & 0.01                                                                        & \underline{0.04}                                                                & \textbf{0.43}                                                              & 0.44                                                                          \\
100 UK + \hspace{0.5ex} 0 HK  & 185                                              & 0.20                                                                       & 0.08                                                                        & \underline{0.36}                                                                & \textbf{0.69}                                                              & 0.77                                                                           & 314                                              & 0.05                                                                        & \underline{0.11}                                                                & \textbf{0.51}                                                              & 0.54                                                                          \\
\hspace{7ex} + \hspace{0.5ex} 1 HK  & 264                                              & 0.14                                                                       & 0.00                                                                        & \underline{0.39}                                                                & \textbf{0.73}                                                              & 0.78                                                                           & 425                                              & 0.00                                                                        & \underline{0.08}                                                                & \textbf{0.53}                                                              & 0.55                                                                          \\
\hspace{7ex} + 10 HK & 213                                              & 0.12                                                                       & 0.01                                   & \underline{0.41}                                                                & \textbf{0.73}                                                              & 0.79                                                                           & 618                                              & 0.01                                                                        & \underline{0.06}                                                                & \textbf{0.49}                                                              & 0.52                                                                          \\
500 UK + \hspace{0.5ex} 1 HK  & 748                                              & 0.12                                                                       & 0.07                                                                        & \underline{0.79}                                                                & \textbf{0.84}                                                              & 0.95                                                                           & 802                                              & 0.01                                                                        & \underline{0.23}                                                                & \textbf{0.81}                                                              & 0.85                                                                          \\
\hspace{7ex} + 10 HK & 568                                              & 0.11                                                                       & 0.01                                        & \underline{0.84}                                                                & \textbf{0.86}                                                              & 0.97                                                                           & 1,134                                             & 0.01                                                                        & \underline{0.22}                                                                & \textbf{0.80}                                                              & 0.83                                                                          \\ \bottomrule
\end{tabular}
}
\caption{\textbf{Reasons for knowledge shifts.} UK $\to$ HK and HK $\to$ UK indicate absolute amount of shifts. Each reason reflects the relative contribution to understanding the nature of shifts. Bold numbers reflect a main reason of shift, underlined numbers - the second best reason of shift.  } 

\label{shifts_explained}
\end{table*}

Table~\ref{shifts_explained} summarizes the absolute amount of shifts from \texttt{Unknown} to \texttt{HighlyKnown} (\textit{UK $\to$ HK}) and from \texttt{HighlyKnown} to \texttt{Unknown} (\textit{HK $\to$ UK}). The numbers for the shifts are relative and reflect the percentage of cases that fall into this or that reason. The reasons we consider are as follows:

\begin{enumerate}
    \item \textit{non-refusal} -- the percentage of cases where the default model refused to answer while the trained model produced correct answer

    \vspace{-1.5ex} 
    
    \item \textit{explosion} --  the percentage of cases where the model predicted one of the exploded answers. The exploded answer is the one to which the model converges in percentage of cases considerably higher than expected by the default model
    
    \vspace{-1.5ex} 

    \item \textit{target-based} -- the percentage of cases where the model predicts one of the answers from the unknown examples it was trained on
    
    \vspace{-1.5ex} 
    
    \item \textit{domain shift} -- all questions in the dataset were constructed by templates in a form of \texttt{<subject, relation, object>}. %Naturally, the relation is associated with the ``sense'' of the question.
    %For example, the question ``1896 FA Cup Final is in which city?'' is constructed from the triple \texttt{<1896 FA Cup Final, city, London>}.
    Each relation in DBpedia has either its range or domain. Both give insight into the way the relation links a subject to an object. In case of domain, the subject qualifies as a type of thing specified in the domain. The range works exactly like the domain but applies to the object. 390 relations from the test data fall into 92 relation domain categories. For example, the relation domain ``PopulatedPlace'' includes 24 relations, allowing us to analyze shifts of knowledge of a large amount of similar relations instead of looking at the specific one. \textit{domain shift} shows the percentage of cases where the shift goes from the same domain or range of the relation for the cases the model was trained on
    
    \vspace{-1.5ex} 

    \item \textit{shift explained} -- the percentage of shift explained by all aforementioned reasons. Note that the sum of the rates for all reasons does not necessarily sum to the percentage of the shift explained, since some of the reasons overlap. 
    % For example, there are examples which fall into categories ``from exploded'' and ``from target'' simultaneously
    
\end{enumerate}

Except for the \textit{non-refusal} all reasons fall into both positive and negative shifts. For example, we see that both positive and negative shifts occur inside the domain. Specific examples of positive and negative shifts are presented in Appendix~\ref{sec:examples_shifts}.

\begin{figure}[ht!]
    \centering
    \includegraphics[scale=0.55]{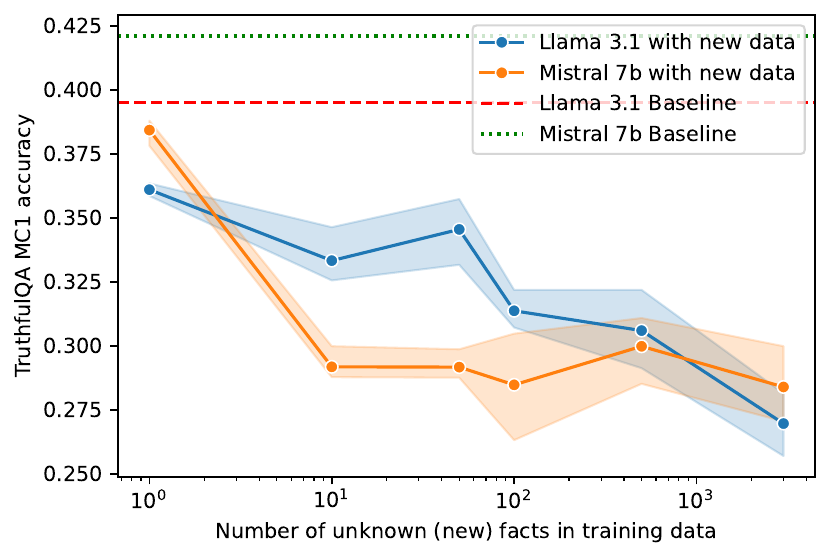}
    
    \caption{Difference in quality drop for the two models: Llama 3.1 8B and Mistral 7B v0.3.}
    \label{fig:mistral_add_main_example}
\end{figure}

\paragraph{Discussion}

If we train the model just for 1 new unknown fact, the positive shift occurs mainly because the model starts answering questions it has not answered previously. 

From the exploded answers suffer the models (10 UK+ 1 HK and 10 UK + 0 HK) that have a low amount of training data with a large proportion of new trained unknown facts relative to the size of the training data.

Finally, we can see that with the increase in the amount of learned unknown facts, there is an increase of the rate of domain shift and target spillover for positive shifts. On the other hand, for the negative shifts, there is only the tendency for the increase of the rate of the domain shift explaining the proportion of the shift explained. Besides, if we compare target-based percentages for positive and negative shifts, we can see that for all models this percentage is higher for positive shifts. The same is true for the negative shifts and for the domain shift as well. It means that out of these two reasons the positive effect of learning from the same domain and target is higher than negative.

\section{Additional Results for Mistral}

Similar effects to those described above for the LLaMa model may be observed in other models, such as GPT-3 and Mistral. %We tested Mistral-7B-Instruct v0.3\footnote{\url{https://hf.co/mistralai/Mistral-7B-Instruct-v0.3}}.
We have conducted a set of experiments for the \textsf{\small{Mistral-7B-Instruct-v0.3}} model, to verify if the results obtained for the \textsf{\small{Llama-3.1-8B-Instruct}} model are generalizable to other decoder-only models. 

As we can see in Table~\ref{table:mistral_add_main_example}, the number of Paraphrases and \texttt{HighlyKnown} increases, so does the positive shift. This corroborates our findings on \textsf{\small{Llama-3.1-8B-Instruct.}}

\begin{table}[!ht]
    \centering
    \footnotesize
    \setlength{\tabcolsep}{2pt}
\begin{tabular}{lccccccc}
\toprule

\multirow{2}{*}{\textbf{Method}}            & \multicolumn{2}{c}{\textbf{Accuracy shift}}\\ %{\textbf{Accuracy Shift}} & { \textbf{}} & {\textbf{}}\\
{}                             & {Pos. shift}     & { Neg. shift}             \\ \midrule
{1 Unknown + 1 Paraphrase}          & {0.159}     & {0.106}           \\
{1 Unknown + 10 Paraphrases}         & {0.196}    & {0.114}          \\
% {1 Unknown + 10 Paraphrase}        & {0.5546}    & {0.3164}          \\ 
\midrule
                                     &     &             \\ 
\midrule
{1 Unknown + 1 HighlyKnown}        & {0.160}    & {0.114}          \\
{1 Unknown + 10 HighlyKnown}        & {0.174}     & {0.146}          \\
%{10 Unknown + 10 Paraphrase}        & {0.5299}    & {0.3088}          \\ \midrule
%                                    &       &                \\ 
%\midrule
%{1 Unknown + 0 HighKnown}    & {0.7967}       & {0.5520}     & {0.3210}           \\
%{1 Unknown + 1 HighKnown}    & {0.7988}       & {0.5538}    & {0.3149}          \\
%{1 Unknown + 10 HighKnown}   & {0.7723}       & %{0.5333}    & {0.3287}          \\ \hline
%                          &             &     &             \\ 
%\midrule
%{10 Unknown + 0 HighKnown}   & {0.7252}       & {0.5213}    & {0.3272}          \\
%{10 Unknown + 1 HighKnown}   & {0.7218}       & {0.5247}    & {0.3041}          \\
%{10 Unknown + 10 HighKnown}  & {0.7517}       & {0.5324}    & {0.2980}           \\ 

\bottomrule
\end{tabular}                   
\caption{Accuracy shift for Mistral-7B-Instruct-v0.3.}
\label{table:mistral_add_main_example}

\end{table}

As can be seen in Figure \ref{fig:mistral_add_main_example}, the quality of the model decreases as the number of new knowledge increases. Models are not strongly correlated, but we see similar features, such as a significant drop from 1 new knowledge to 10, and a small recovery of quality at 100 new knowledge. Thus, we believe that generalizing our research to other decoder-only models is possible.

\section{Conclusion}

Our findings revealed that the most significant increase in the acquisition of additional knowledge is achieved when a blend of \texttt{Unknown} and \texttt{HighlyKnown} data is incorporated into the training phase. However, this approach comes with a trade-off: it compromises the model's capability to accurately answer complex or nuanced questions, as assessed by TruthfulQA.

Additionally, it has been observed that the incorporation of a limited number of unknown facts within a vast set of \texttt{HighlyKnown} examples or paraphrases  can substantially impair the model's reasoning abilities as measured by MMLU.

In an additional observation, it was noted that upon being fine-tuned with LoRA adapters, LLMs experience a reduction in their ability to express uncertainty when formulating answers. In certain scenarios, this leads to the models disproportionately favoring responses that are statistically overrepresented.

\section*{Acknowledgement}

This research was conducted within the framework of a joint MTS-Skoltech laboratory on AI.

\section*{Limitations}

Since we only used one LLM for our study (namely, \textsf{\small{Llama-3.1-8B-Instruct}}), it's unknown if the outcomes will change when utilizing different LLMs. However, because our work requires a lot of computation, it is difficult to repeat on several LLMs. First, we experimented a lot with different learning rates, LoRA ranks and used two data augmentation techniques with 0, 1, 10 extra examples per sample. This results in a great number of experiments. Second, and perhaps most crucially, we had to annotate a sizable dataset with the knowledge categories to make our analysis easier. It was essential to precisely evaluate the model's knowledge with regard to a particular fine-tuning case in order to draw trustworthy findings. Beyond this, detecting knowledge categories is a machine-intensive task not only for defining the initial categories, but at each inference step. 

In addition, there are a few crucial directions for future research left unexplored in this paper. Namely, how early stopping could have influenced the distribution of knowledge categories and models' behavior on external benchmarks. Adding training examples from the same relation domain or range as well as few-shot prompts from this category could have also boosted the performance. We leave these questions for further exploration.

\section*{Ethical Considerations}

We have carefully curated the generated dataset, and we have not encountered any inappropriate or offensive content within it.

\bibliography{custom}
\onecolumn 
\appendix

\section{MMLU Performance}
\label{sec:appendix_mmlu}

\begin{table*}[!ht]
    \centering
    \footnotesize
    \setlength{\tabcolsep}{2pt}
\begin{tabular}{l|cc|cccccc}
\toprule
                          & \multicolumn{2}{c|}{{\underline{ MMLU}}}                                            & \multicolumn{6}{c}{{\underline{ ThruthfulQA}}}                                                                                                                                                                                                \\ \midrule
                          & \multicolumn{1}{l}{\textbf{Accuracy}} & \multicolumn{1}{l|}{\textbf{Std}} & \multicolumn{1}{l}{\textbf{BLEU}} & \multicolumn{1}{l}{\textbf{ROUGE-1}} & \multicolumn{1}{l}{\textbf{ROUGE-2}} & \multicolumn{1}{l}{\textbf{ROUGE-L}} & \multicolumn{1}{l}{\textbf{MC1 Acc.}} & \multicolumn{1}{l}{\textbf{MC2 Acc.}} \\ \midrule
Llama-3.1-8B-Instruct     & 0.677                                 & 0.004                                & 35.780                            & 60.325                               & 47.789                               & 58.512                               & 0.382                                & 0.552                                \\ \midrule
\hspace{2.5ex} 1 UK\hspace{1.3ex} + \hspace{3.5ex} 0 HK       & 0.682                                 & 0.004                                & 33.741                            & 57.414                               & 44.176                               & 55.374                               & 0.359                                & 0.533                                \\
\hspace{9.3ex} + \hspace{3.5ex} 1 HK       & 0.681                                 & 0.004                                & 34.920                            & 58.537                               & 45.814                               & 56.587                               & 0.361                                & 0.538                                \\
 \hspace{9.3ex} + \hspace{2.5ex} 10 HK    & 0.672                                 & 0.004                                & 25.367                            & 44.604                               & 30.873                               & 42.816                               & 0.322                                & 0.478                                \\
\hspace{1.5ex} 10 UK\hspace{1.2ex} + \hspace{3.5ex} 0 HK       & 0.615                                 & 0.004                                & 12.524                            & 22.778                               & 11.987                               & 21.519                              & 0.305                              & 0.481                              \\
\hspace{9.3ex} + \hspace{3.5ex} 1 HK      & 0.627                                 & 0.004                                & 13.251                            & 27.477                               & 15.078                               & 25.530                               & 0.310                                & 0.451                                \\
 \hspace{9.3ex} + \hspace{2.5ex} 10 HK     & 0.653                                 & 0.004                                & 21.659                            & 39.459                               & 25.678                               & 37.562                               & 0.319                                & 0.490                                \\
\hspace{1.5ex} 50 UK\hspace{1.2ex} + \hspace{3.5ex} 0 HK     & 0.660                              & 0.004                                & 18.906                           & 34.491                            & 19.259                               & 32.569                               & 0.306                               & 0.482                                \\
\hspace{9.3ex} + \hspace{2.5ex} 1 HK     & 0.675                                 & 0.004                                & 18.740                            & 35.436                               & 19.318                               & 33.398                               & 0.313                                & 0.494                                \\
 \hspace{9.3ex} + \hspace{2.5ex} 10 HK   & 0.658                                 & 0.004                                & 5.592                             & 20.859                               & 9.420                                & 19.321                               & 0.282                                & 0.448                                \\
\hspace{1.1ex} 100 UK\hspace{0.5ex} + \hspace{3.5ex} 0 HK     & 0.663                                & 0.004                                & 16.987                         & 34.884                            & 19.598                               & 32.935                               & 0.304                                & 0.474                             \\
\hspace{9.3ex} + \hspace{3.5ex} 1 HK    & 0.669                                 & 0.004                                & 19.586                            & 37.386                               & 22.375                               & 35.500                               & 0.315                                & 0.482                                \\
 \hspace{9.3ex} + \hspace{2.5ex} 10 HK   & 0.659                                 & 0.004                                & 9.609                             & 27.236                               & 13.829                               & 25.363                               & 0.277                                & 0.452                                \\
\hspace{1.1ex} 500 UK\hspace{0.5ex} + \hspace{3.5ex} 0 HK      &   0.649                                    &                            0.004          &     10.115                              &      25.991                                &      12.541                                &                          23.909            &  0.290                                    &    0.447                                  \\
\hspace{9.3ex} + \hspace{3.5ex} 1 HK     & 0.655                                 & 0.004                                & 7.561                             & 23.507                               & 11.480                               & 21.498                               & 0.297                                & 0.460                                \\
 \hspace{9.3ex} + \hspace{2.5ex} 10 HK  & 0.554                                 & 0.004                                & 6.829                             & 23.143                               & 9.583                                & 21.519                               & 0.296                                & 0.463                                \\
3000 UK \hspace{0.5ex} + \hspace{3.5ex} 0 HK   &                           0.579            &   0.004                                   &    11.415                               &        27.783                              &                        14.884              &          25.363                            &     0.294                                 &   0.461                                   \\
\hspace{9.3ex} + \hspace{3.5ex} 1 HK   &  0.623                                     &          0.004                            &      5.561                             &                       19.906               &    7.422                                  &    18.280                                  &         0.257                             &                     0.420                 \\
 \hspace{9.3ex} + \hspace{2.5ex} 10 HK   &   0.554 &         0.004    &          9.239  &          23.447     &           11.558     &    21.415      &         0.263      &      0.445     \\ \midrule
\hspace{2.5ex} 1 UK\hspace{1.3ex} + \hspace{1.5ex} 0 Paraphrase      &  0.682              &  0.004             &           33.741   &      57.414          &   44.176        &          55.374       &       0.359          &              0.533   \\
 \hspace{9.3ex} + \hspace{1.5ex} 1 Paraphrase    & 0.681                                 & 0.004                                & 36.582                            & 60.991                               & 48.642                               & 59.052                               & 0.365                                & 0.537                                \\
 \hspace{9.3ex} + \hspace{0.5ex} 10 Paraphrase    & 0.674                                 & 0.004                                & 35.930                            & 59.606                               & 46.787                               & 57.851                               & 0.356                                & 0.535                                \\
\hspace{1.5ex} 10 UK\hspace{1.2ex} + \hspace{1.5ex} 0 Paraphrase    & 0.615                                 & 0.004                                & 12.524                            & 22.778                               & 11.987                               & 21.519                               & 0.305                                & 0.481                                \\
 \hspace{9.3ex} + \hspace{1.5ex} 1 Paraphrase     & 0.610                                 & 0.004                                & 18.518                            & 36.578                               & 23.061                               & 34.481                               & 0.343                                & 0.527                                \\
 \hspace{9.3ex} + \hspace{0.5ex} 10 Paraphrase  & 0.624                                 & 0.004                                & 15.420                            & 31.984                               & 20.046                               & 29.987                               & 0.332                                & 0.495                                \\
\hspace{1.5ex} 50 UK\hspace{1.2ex} + \hspace{1.5ex} 0 Paraphrase    & 0.660                                 & 0.004                                & 18.906                            & 34.491                               & 19.259                               & 32.569                               & 0.306                                & 0.482                                \\
 \hspace{9.3ex} + \hspace{1.5ex} 1 Paraphrase   & 0.670                                 & 0.004                                & 17.795                            & 37.690                               & 21.159                               & 35.887                               & 0.344                                & 0.517                                \\
 \hspace{9.3ex} + \hspace{0.5ex} 10 Paraphrase  & 0.668                                 & 0.004                                & 17.944                            & 38.240                               & 22.340                               & 35.891                               & 0.337                                & 0.511                                \\
\hspace{0.5ex} 100 UK\hspace{1.2ex}  + \hspace{1.5ex} 0 Paraphrase    & 0.663                                 & 0.004                                & 16.987                            & 34.884                               & 19.598                               & 32.935                               & 0.304                                & 0.474                                \\
 \hspace{9.3ex} + \hspace{1.5ex} 1 Paraphrase  & 0.675                                 & 0.004                                & 19.882                            & 40.465                               & 24.316                               & 38.299                               & 0.334                                & 0.507                                \\
 \hspace{9.3ex} + \hspace{0.5ex} 10 Paraphrase  & 0.659                                 & 0.004                                & 25.621                            & 47.628                               & 32.787                               & 45.445                               & 0.337                                & 0.511                                \\
\hspace{0.5ex} 500 UK\hspace{1.2ex} + \hspace{1.5ex} 0 Paraphrase    & 0.649                                 & 0.004                                & 10.115                            & 25.991                               & 12.541                               & 23.909                               & 0.290                                & 0.447                                \\
 \hspace{9.3ex} + \hspace{1.5ex} 1 Paraphrase  &  0.653                                     &      0.004                                &      14.245                             &         33.977                             &                            17.747          &     31.160                                 &      0.322                                &       0.472                               \\
 \hspace{9.3ex} + \hspace{0.5ex} 10 Paraphrase  & 0.566                                 & 0.004                                & 10.514                            & 26.604                               & 12.961                               & 24.337                               & 0.343                                & 0.512                                \\
3000 UK \hspace{0.5ex} + \hspace{1.5ex} 0 Paraphrase  & 0.579                                 & 0.004                                & 11.415                            & 27.783                               & 14.884                               & 25.363                               & 0.294                                & 0.461                                \\
\hspace{9.3ex} + \hspace{1.5ex} 1 Paraphrase  & 0.605                                 & 0.004                                & 17.107                            & 38.457                               & 22.222                               & 35.300                               & 0.319                                & 0.502                                \\
 \hspace{9.3ex} + \hspace{0.5ex} 10 Paraphrase & 0.441                                 & 0.004                                & 29.352                            & 50.871                               & 40.383                               & 49.400                               & 0.305                                & 0.491                                \\ \bottomrule
\end{tabular}                   
\caption{Accuracy for MMLU and a range of metrics  for ThruthfulQA for all trained LoRA adapters}
\label{table:benchmarks}
\end{table*}

\section{Few-shot Prompts}

\label{sec:few-shot}

Here is an example of the 4-shot prompt. The prompt includes 4 question-answer examples, 5th question is the question of interest.

\begin{tcolorbox}[colback=gray!3, colframe=gray!50,fontupper=\itshape]

Answer the following question. 

Question: Who wrote the novel Evening Class? Answer: maeve binchy

Question: Which country does the airline Air Pacific come from? Answer: fidji 

Question: In which branch of the arts does Allegra Kent work? Answer: balletti

Question: Who had a 70s No 1 hit with Billy, Don't Be A Hero? Answer: bo donaldson and heywoods

Question: 12th Brigade (Australia) fought in what battle?
\end{tcolorbox}

\section{Samples of Knowledge Categories}

\label{sec:examples_categories}

\begin{tcolorbox}[colback=gray!3, colframe=gray!50]

\texttt{HighlyKnown.} \textit{Question: where is Alfa Romeo MiTo assembled? Answer: Turin}

\texttt{MaybeKnown.} \textit{Question: Daredevil (TV series) is on which channel? Answer: Netflix}

\texttt{Unknown.} \textit{Question: Can you name a band member of Ashes of Ares? Answer: Matthew Barlow}

\end{tcolorbox}

% \begin{tcolorbox}[colback=gray!3, colframe=gray!50]

% \texttt{Initial question.} \textit{Who was a notable student of Arthur Eddington?}

% \texttt{Paraphrase.} \textit{What student of Arthur Eddington made significant contributions to the field?}

% \texttt{HighlyKnown.} \textit{Where is Alfa Romeo MiTo assembled?}

% \end{tcolorbox}

\section{Examples of Positive and Negative Shifts}

\label{sec:examples_shifts}

% Let’s look at the case study of 1UK+10HK. Before training the model answered to the question: “Aalen is in which administrative district?” "Norway", and after training it produced the correct answer - "Stuttgart Government Region".

% Examples of \textbf{positive shifts}, the case when the model starts to answer correctly the questions for which it has not been specifically trained (in the answer would come first the answer before training and second the answer after training):
Let's illustrate the concept of positive shifts through examples.
This occurs when a model begins to accurately respond to queries for which it has not been specifically trained (A\textsubscript{before} -- an answer before fine-tuning, A\textsubscript{after} -- an answer after fine-tuning).\\
\textbf{Positive Shifts:}
\begin{tcolorbox}[colback=gray!3, colframe=gray!50]

Q: \textit{Batata vada is located in which region?} A\textsubscript{before}: \textit{India} A\textsubscript{after}: \textit{Maharashtra} 
% Explanation: same domain as in the trained question

Q: \textit{Esslingen am Neckar is in which administrative?} A\textsubscript{before}: \textit{Esslingen} A\textsubscript{after}: \textit{Stuttgart Government Region} 
% (\textbf{from target}- the same answer as in target) 

Q: \textit{What is the home arena of Anaheim Storm?} A\textsubscript{before}: \textit{I couldn't find any information} A\textsubscript{after}: \textit{Anaheim Arena}
% (\textbf{non-refusal}: model starts to answer the question it refused to answer before training)

\end{tcolorbox}

\textbf{Negative Shifts:}

\begin{tcolorbox}[colback=gray!3, colframe=gray!50]

Q: \textit{Amir Mokri was born in which place?} A\textsubscript{before}: \textit{Iran} A\textsubscript{after}: \textit{Tehran}
% (\textbf{exploded answer}: the model converges to answer “London” too often) 

Q: \textit{In which canton is Gachnang located?} A\textsubscript{before}: \textit{Thurgovia} A\textsubscript{after}: \textit{aargau} 
% (\textbf{negative domain shift} - the model shifts to some representations of the toponyms incorrectly)
\end{tcolorbox}

\section{ARC and LogiQA Benchmarks}
\label{sec:extra_benchmarks}

\begin{table*}[!ht]
    \centering
    \footnotesize
    \setlength{\tabcolsep}{2pt}
\begin{tabular}{lcccccccc}
\toprule

{\textbf{Method}}            & \multicolumn{2}{c}{{\textbf{ARC}} } & {\textbf{LogiQA}} \\
{Llama-3.1-8B-Instruct}                      & { ARC-E}        & {ARC-C}     & { Acc}             \\ \midrule
{1 Unknown + 0 Paraphrase}   & {0.7967}       & {0.5520}     & {0.3210}           \\
{1 Unknown + 1 Paraphrase}   & {0.7942}       & {0.5512}    & {0.3226}          \\
{1 Unknown + 10 Paraphrase} & {0.7875}       & {0.5546}    & {0.3164}          \\ 
\midrule
                        &             &     &             \\ 
\midrule
{10 Unknown + 0 Paraphrase}  & {0.7252}       & {0.5213}    & {0.3272}          \\
{10 Unknown + 1 Paraphrase}  & {0.7386}       & {0.5350}     & {0.3149}          \\
{10 Unknown + 10 Paraphrase} & {0.7142}       & {0.5299}    & {0.3088}          \\ \midrule
                          &          &       &                \\ 
\midrule
{1 Unknown + 0 HighKnown}    & {0.7967}       & {0.5520}     & {0.3210}           \\
{1 Unknown + 1 HighKnown}    & {0.7988}       & {0.5538}    & {0.3149}          \\
{1 Unknown + 10 HighKnown}   & {0.7723}       & {0.5333}    & {0.3287}          \\ \hline
                          &             &     &             \\ 
\midrule
{10 Unknown + 0 HighKnown}   & {0.7252}       & {0.5213}    & {0.3272}          \\
{10 Unknown + 1 HighKnown}   & {0.7218}       & {0.5247}    & {0.3041}          \\
{10 Unknown + 10 HighKnown}  & {0.7517}       & {0.5324}    & {0.2980}           \\

\bottomrule
\end{tabular}                   
\caption{Metrics for ARC and LogiQA benchmarks for  trained LoRA adapters}
\label{extra_benchmarks}

\end{table*}
\end{document}